\documentclass[acmsmall, authoryear, preprint, nonacm]{acmart}

\usepackage{csvsimple}
\usepackage{siunitx}
\usepackage{booktabs}

\begin{document}

\title{Quasi-Monte Carlo Initialization for Meta-Reinforcement Learning}

\author{Julian G. Soltes}
\email{jsoltes@regis.edu}
\orcid{0009-0003-4226-1778}
\affiliation{
  \institution{Regis University}
  \city{Denver}
  \state{Colorado}
  \country{USA}
}

\begin{abstract}
This paper explores the efficacy of quasi-Monte Carlo (QMC) weight initialization for meta-reinforcement learning within modern benchmark environments. Various sampling methods are used to bound a population-based search and aggregate an optimal prior from a baseline set of tasks. The QMC meta-priors $\theta_i^*$ show improvements in training convergence compared to modern orthogonal (SB3) defaults when extrapolated to similar unseen continuous control environments. In dissimilar tasks, the orthogonal orientation was globally superior for an unbiased search.
\end{abstract}

\maketitle
\pagestyle{plain}

\section{Introduction}
\begin{figure}[h]
\includegraphics[width=0.65\textwidth]{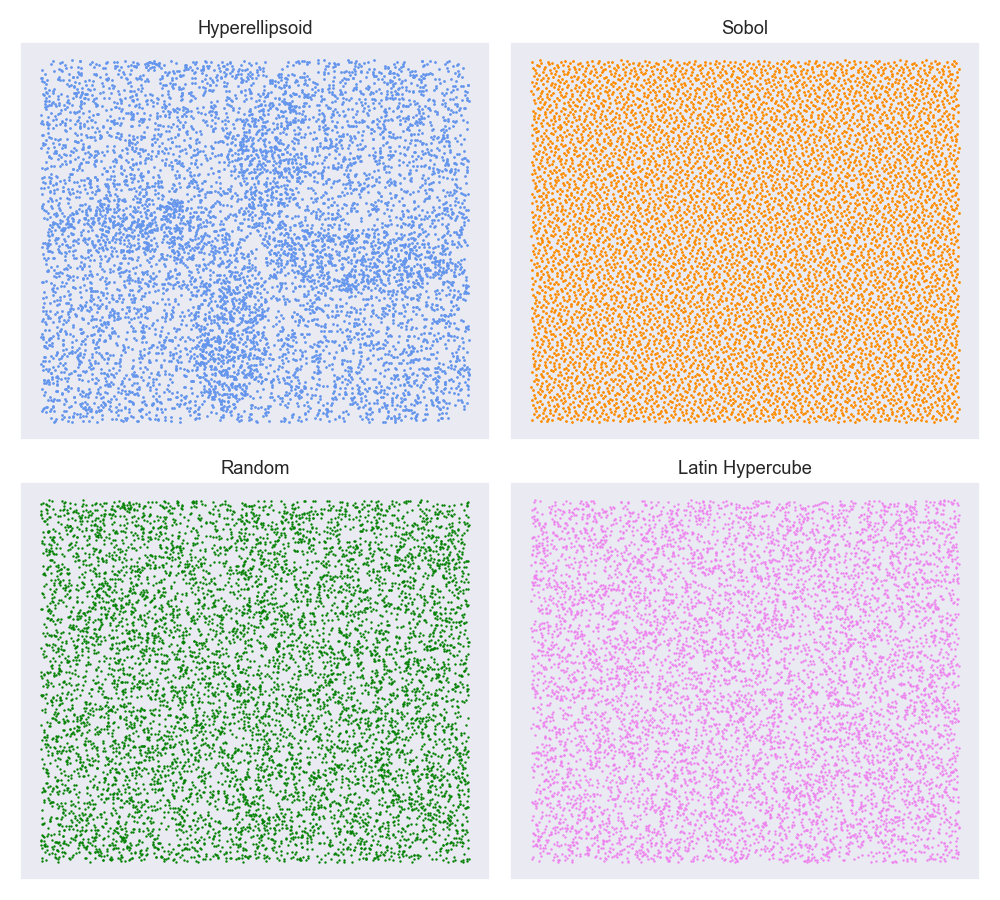}
\caption{Representative 2-D sample populations.}
\label{fig:samples}
\end{figure}

Reinforcement learning (RL) is highly influenced by the model's initial conditions \cite{kwok2025butterfly}. This study uses various quasi-Monte Carlo (QMC) sampling methods $i$ (Fig. \ref{fig:samples}): Sobol, Latin Hypercube (LHS), and Hyperellipsoid Density Sampling (HDS) \cite{soltes2026hyperellipsoiddensitysamplingexploitative}, to create both Euclidean and non-Euclidean search spaces $\Omega_i$ from which to sample initial model weights $\theta_i \in \Omega_i$. An efficient one-step training sweep through this sample population identifies an optimal meta-prior $\theta^*_i$ to accelerate learning when applied to similar unseen environments.

Quasi-Monte Carlo sequences are commonly explored for one-shot optimization problems \cite{10.1007/978-3-030-58112-1_8}. Extrapolating to meta-reinforcement learning, the results of this study show that the exploitative HDS, low-discrepancy Sobol, and high-stratification LHS meta-priors ($\theta_H^*$, $\theta^*_S$, $\theta^*_L$) improve training performance when initializing unseen environments with similar kinematic profiles.

\section{Methodology}
The experimental setup is configured using modern RL architectures (section \ref{sec:setup}). The optimal meta-prior ($\theta_i^*$) is identified and evaluated via one-step adaptation (\ref{sec:adaptation}) and zero-shot transfer (\ref{sec:transfer}).

\subsection{Experimental Setup}
\label{sec:setup}
The experimental trials are conducted using proximal policy optimization (PPO) via \texttt{StableBaseline3} (SB3) \cite{stable-baselines3} on various \texttt{Gymnasium} \cite{towers2024gymnasium} environments. Quasi-Monte Carlo samples were generated using \texttt{hdim\_opt} v1.4.71. All trials are bootstrapped over 16 randomly-seeded iterations to achieve statistically significant results ($p < 0.05$) via the rank-sum test. Each configuration's random seed is defined relative to the experimental trial iteration to ensure reproducibility.

\subsubsection{RL Architecture}
The actor and critic networks are configured as multi-layer perceptrons consisting of two 32-neuron layers to prevent over- or under-fitting to the initial search environments; for generalizability across tasks, the initial sample weights are injected exclusively into these networks, bypassing the environment-specific output layers. The injected sample weights are internally and dynamically scaled via LeCun / Xavier initialization ($\propto1/\sqrt{D}$) \cite{glorot2010understanding}.

The observation spaces are padded to 27 dimensions (the total position and velocity states within Gymnasium's Ant-v5) to account for varying observation shapes between environments. This relatively low dimensionality provides the additional benefit of reducing network overfitting to the one-step search environment(s).

\subsection{One-Step Adaptation}
\label{sec:adaptation}
A weight population $N = 2^{10}$ was generated using HDS, Sobol, LHS, and random baselines (Fig. \ref{fig:samples}). Each sample was injected into the policy and value network torsos and evaluated within three baseline search environments: HalfCheetah-v5, BipedalWalker-v3, and Hopper-v5.

\begin{figure}[h!]
\includegraphics[width=0.55\textwidth]{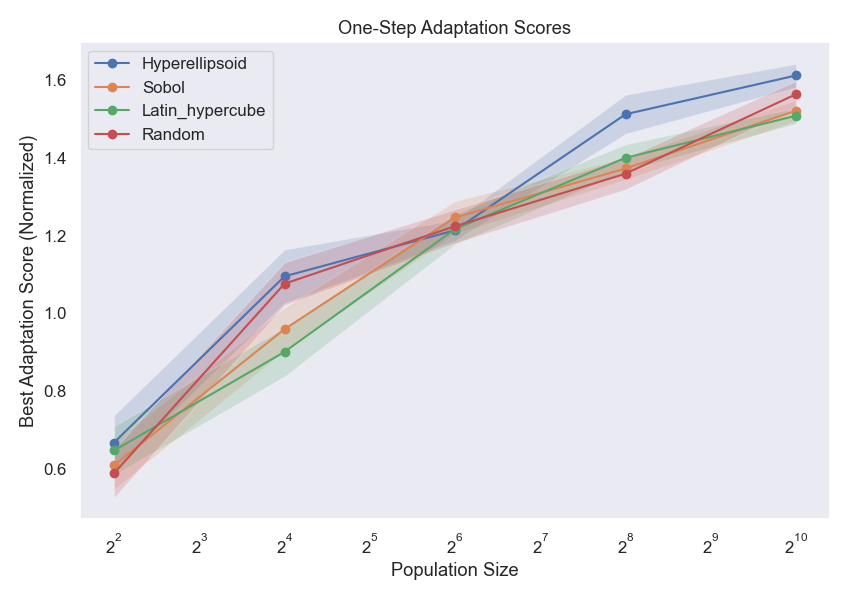}
\caption{Best adaptation scores $\mathcal{A}^+$ using various population sizes.}
\label{fig:initial_adaptation}
\end{figure}

The PPO is trained for a single timestep within each search environment, using Huber loss (smooth $L_1$) and gradient norm clipping ($\vert{}\vert{}g\vert{}\vert{}_2 \leq 1.0$) to stabilize the stochastic gradient descent update with an explorative learning rate of 0.01. The resulting adaptation score $\mathcal{A}$ is quantified as the Euclidean shift in the action distribution mean, Z-score normalized and averaged across the three search environments. The sample weight $\theta_i$ with the highest score $\mathcal{A}^+$ is extracted as the optimal meta-prior $\theta_i^*$ for zero-shot transfer, tested at various sample sizes $N \in \{2^2, \dots, 2^{10}\}$ (Fig. \ref{fig:initial_adaptation}) to visualize $\mathcal{A}^+$ as a function of $N$.

\subsection{Zero-Shot Transfer}
\label{sec:transfer}
The PPO was injected with the optimal meta-prior $\theta_i^*$ and trained for $10^6$ timesteps in three similar (and two dissimilar) unseen continuous control environments: Ant-v5, Walker2d-v5, BipedalWalkerHardcore-v3, Swimmer-v5, and LunarLander-v3. The latter two are included to evaluate the $\theta_i^*$ on tasks with dissimilar profiles to the search environments. The results are compared to training initialized with the SB3 PPO's defaults ('Orthogonal' and 'Random').

\section{Results}
The results confirm the hypothesis that the QMC meta-priors improve performance on environments with similar kinematic profiles, while showing negative effects on dissimilar tasks.

The global converged training performance is aggregated across all evaluation environments and shown in Table \ref{tab:global_performance}, where the $p$-value represents the difference in performance compared to the default SB3 Orthogonal.

\begin{table}[h]
\centering
\caption{Zero-shot transfer performance, Z-normalized across all evaluation environments. $p$-values are calculated relative to the SB3 Orthogonal baseline.}
\vspace{-0.5em}
\label{tab:global_performance}
\begin{tabular}{r|cccc}
\toprule
 & \multicolumn{2}{c}{Similar} & \multicolumn{2}{c}{Dissimilar} \\
\cmidrule(lr){2-3} \cmidrule(lr){4-5}
Strategy & Z ($\mu$ $\pm$ $\sigma$) & $p$ & Z ($\mu$ $\pm$ $\sigma$) & $p$ \\
\midrule
HDS $\theta^*_H$ & $+0.15 \pm 1.04$ & $0.0898$ & $-0.63 \pm 0.77$ & $<0.001$ \\
Sobol $\theta^*_S$ & $+0.20 \pm 0.96$ & $0.0358$ & $-0.22 \pm 0.77$ & $<0.001$ \\
LHS $\theta^*_L$ & $+0.00 \pm 0.94$ & $0.2485$ & $-0.21 \pm 0.66$ & $<0.001$ \\
SB3 Random & $-0.12 \pm 0.88$ & $0.4795$ & $-0.11 \pm 0.71$ & $<0.001$ \\
SB3 Orthogonal & $-0.23 \pm 1.11$ & - & $+1.17 \pm 1.02$ & - \\
\bottomrule
\end{tabular}
\end{table}

\subsection{Similar Tasks}
The similar tasks saw improvements in performance for $\theta_i^*$ compared to SB3 Random and Orthogonal initializations, with varying levels of statistical significance (Table \ref{tab:global_performance}). SB3 Orthogonal performed lowest compared to all methods on Ant-v5; its best task was Walker2d, where it was beat only, and insignificantly, by Sobol $\theta_S^*$ ($p \approx 0.87$). While all QMC meta-priors globally out-performed SB3 Orthogonal, only Sobol $\theta_S^*$ showed statistically significant ($p < 0.05$) improvements.

\begin{figure}[h!]
\includegraphics[width=0.8\textwidth]{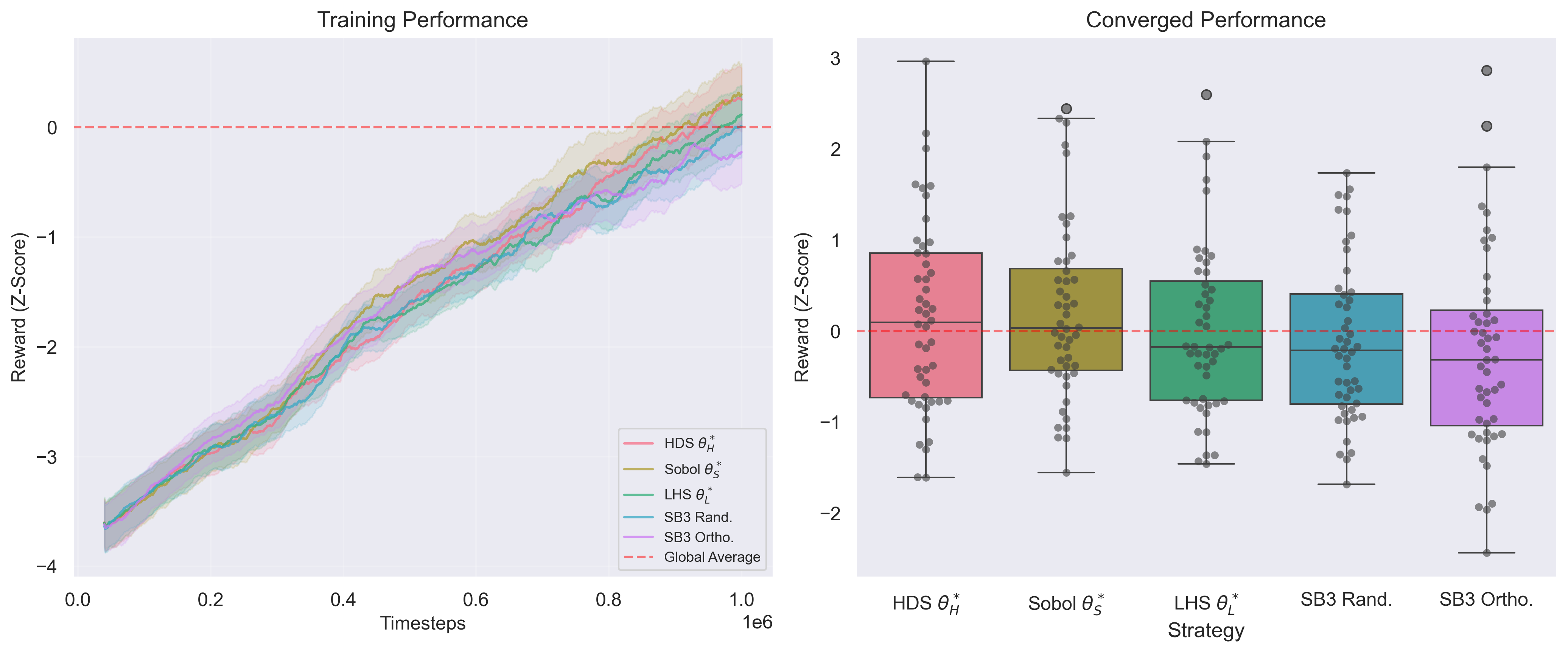}
\caption{Aggregated performance of the meta-priors in similar environments.}
\label{fig:global_similar}
\end{figure}

\subsection{Dissimilar Tasks}
The dissimilar tasks measured global decreases in performance for $\theta_i^*$ compared to SB3 Orthogonal, where the mathematically unbiased initial orientation proved superior within foreign environments ($p<0.001$). The uniform QMC meta-priors (Sobol $\theta_S^*$ and LHS $\theta_L^*$) retain higher performance in these environments compared to the non-Euclidean HDS $\theta_H^*$ due to less overfitting.

\begin{figure}[h]
\includegraphics[width=0.8\textwidth]{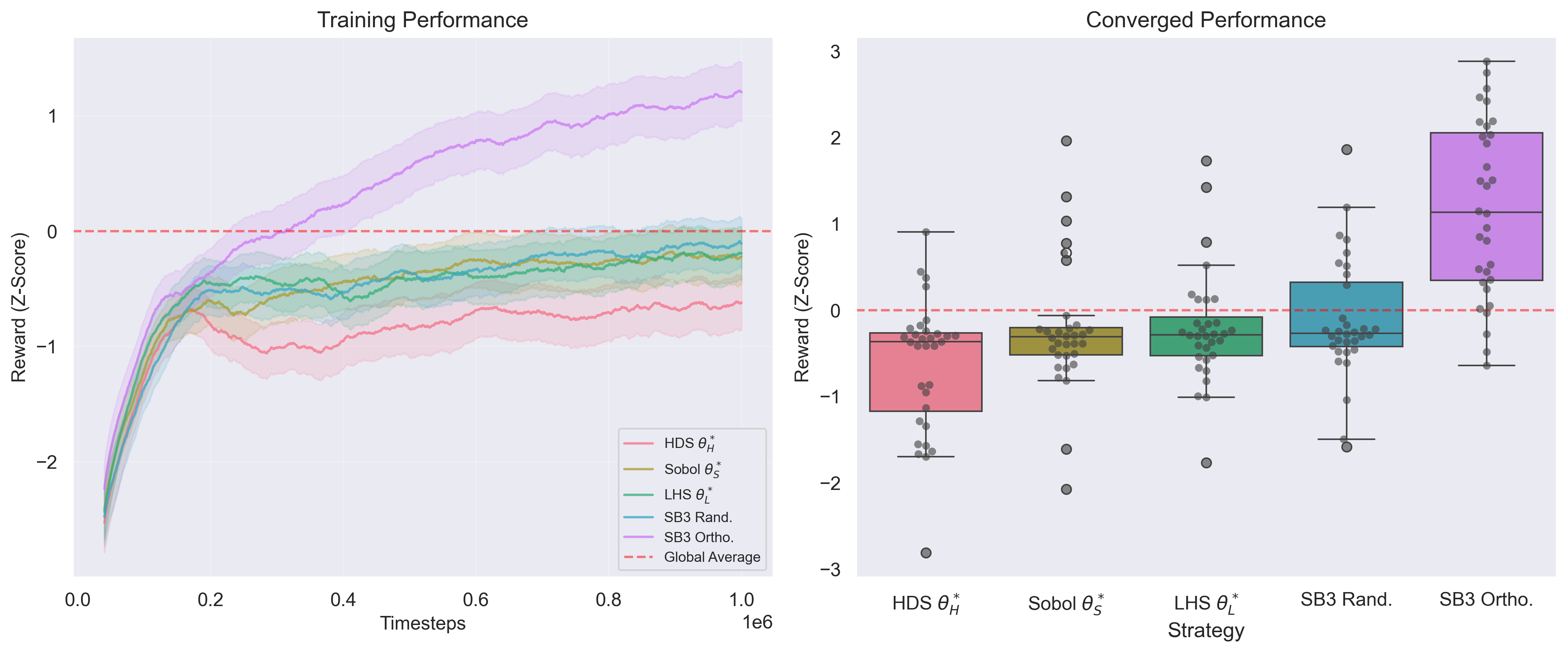}
\caption{Aggregated converged performance of the meta-priors in dissimilar environments.}
\label{fig:global_dissimilar}
\end{figure}

\section{Conclusion}
The results validate quasi-Monte Carlo sampling methods as effective search geometries for zero-shot weight initialization in meta-reinforcement learning problems. The optimal meta-priors $\theta_i^*$ identified from the efficient one-step search are found to accelerate training when initializing similar unseen environments. For dissimilar tasks, the unbiased geometry of SB3 Orthogonal initialization showed globally superior performance.

\begin{acks}
The author would like to thank Dr. Mike Busch for the continued instruction on reinforcement learning architectures. Monet is a source of inspiration, as always.
\end{acks}

\clearpage
\section*{Appendix}
\begin{figure}[h!]
\includegraphics[width=0.8\textwidth]{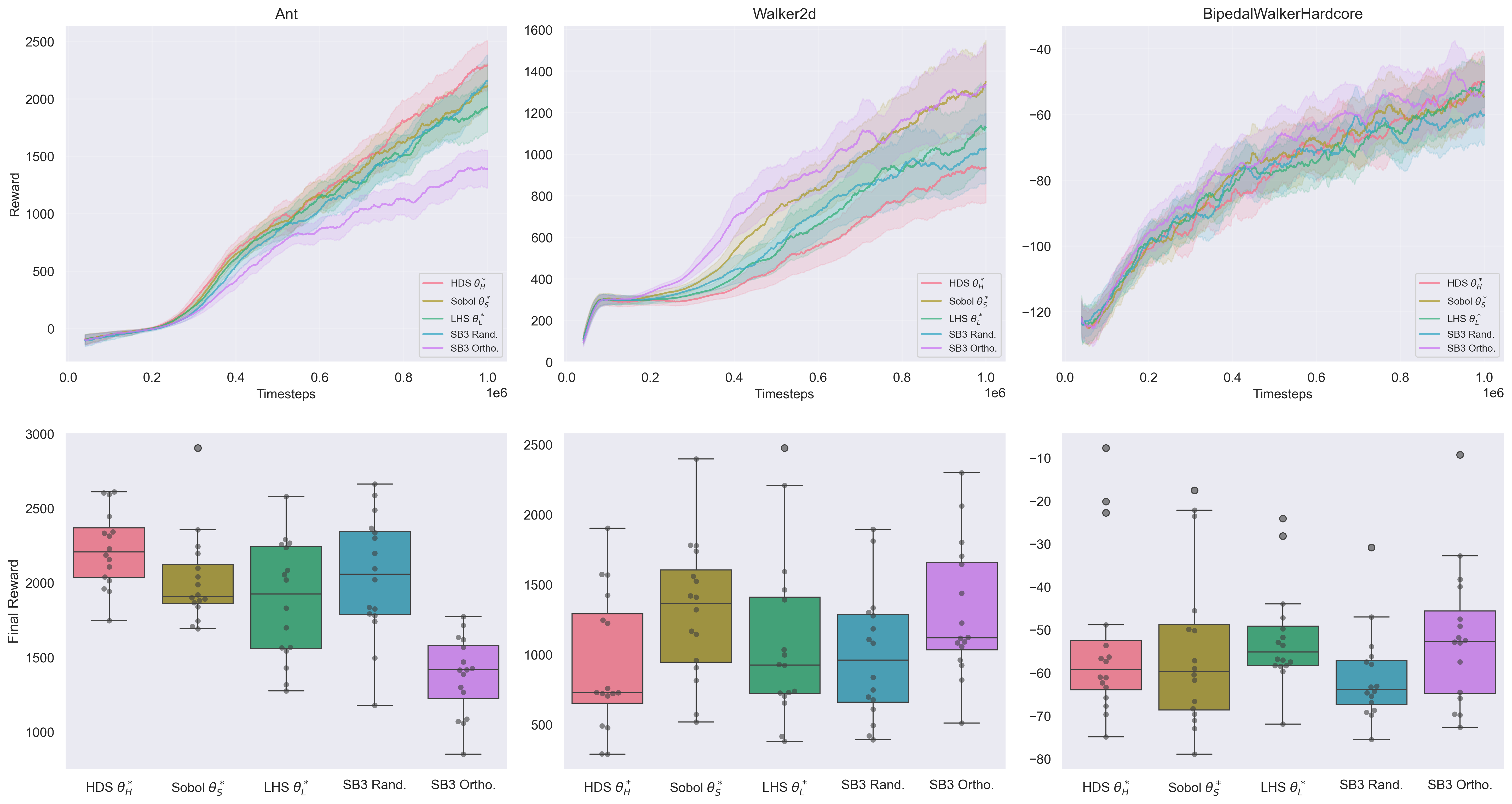}
\caption{Per-environment performance of the meta-priors in similar environments.}
\label{fig:similar}
\end{figure}

\begin{figure}[h!]
\includegraphics[width=0.7\textwidth]{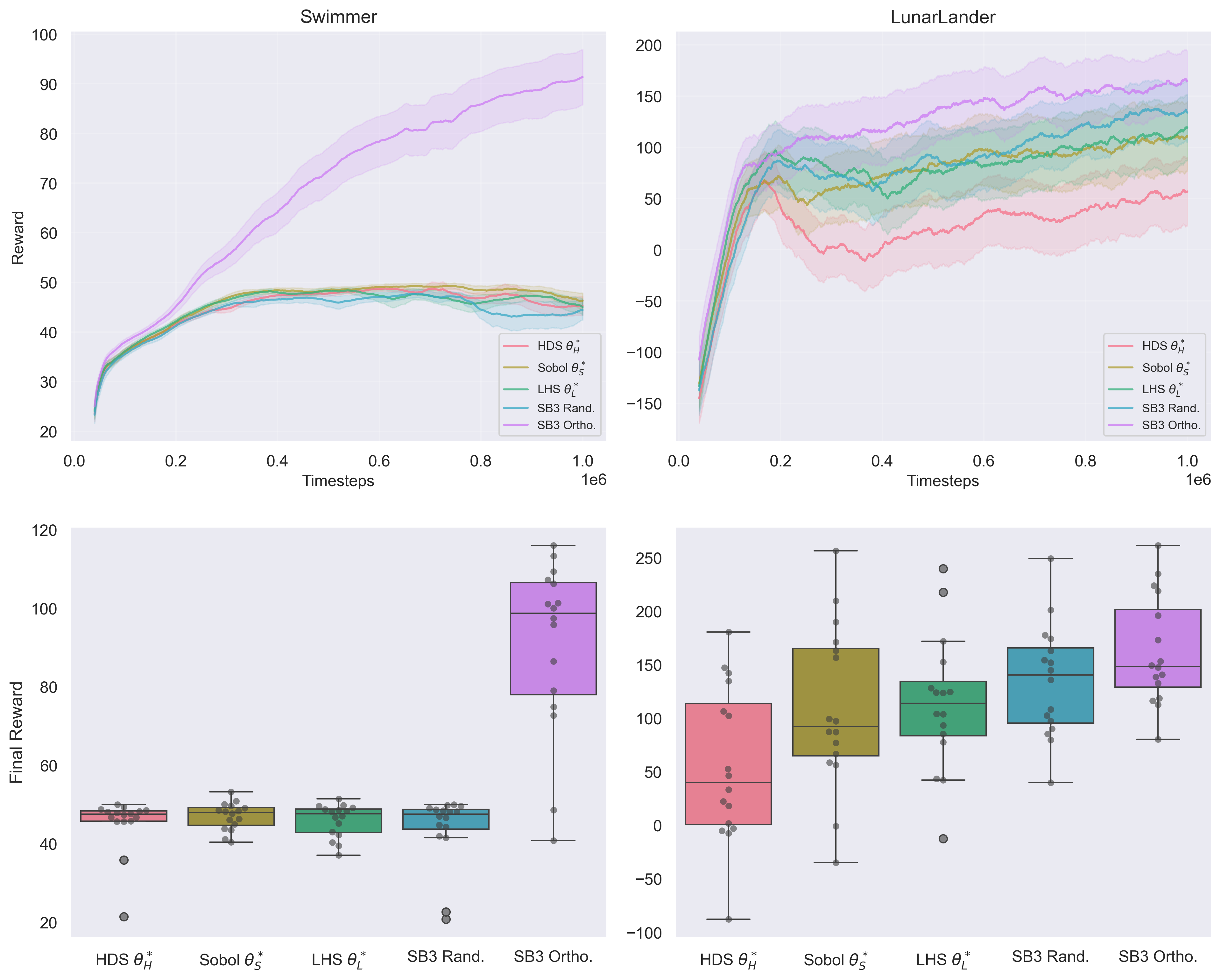}
\caption{Per-environment performance of the meta-priors in dissimilar environments.}
\label{fig:dissimilar}
\end{figure}

\clearpage
\nocite{*}
\bibliographystyle{ACM-Reference-Format}
\bibliography{myreferences.bib}
\end{document}